\documentclass{article}

\usepackage{PRIMEarxiv}

\usepackage[utf8]{inputenc} 
\usepackage[T1]{fontenc}    
\usepackage{hyperref}       
\usepackage{url}            
\usepackage{booktabs}       
\usepackage{amsfonts}       
\usepackage{nicefrac}       
\usepackage{microtype}      
\usepackage{lipsum}
\usepackage{graphicx}
\usepackage{algorithm}
\usepackage[noend]{algpseudocode}
\usepackage{multirow}
\usepackage{fancyhdr}       
\usepackage{graphicx}       
\graphicspath{{media/}}     

\title{Association Rules Mining with Auto-Encoders
}

\author{
  Théophile Berteloot, Richard Khoury, Audrey Durand \\
  Université Laval \\
  Québec City \\
}

\begin{document}
\maketitle

\begin{abstract}
Association rule mining is one of the most studied research fields of data mining, with applications ranging from grocery basket problems to explainable classification systems. Classical association rule mining algorithms have several limitations, especially with regards to their high execution times and number of rules produced. Over the past decade, neural network solutions have been used to solve various optimization problems, such as classification, regression or clustering. However there are  still no efficient way association rules using neural networks. In this paper, we present an auto-encoder solution to mine association rule called ARM-AE. We compare our algorithm to FP-Growth and NSGAII on three categorical datasets, and show that our algorithm discovers high support and confidence rule set and has a better execution time than classical methods while preserving the quality of the rule set produced.
\end{abstract}


\section{Introduction}
\label{introduction}
Association rule mining (ARM) was first introduced by Agrawal \cite{agrawal1994fast} to solve the grocery basket problem, and since then it has found numerous applications in Knowledge Discovery in Database (KDD) problems ranging from financial analysis \cite{nair2015stock} to medical diagnostics \cite{alkeshuosh2017using}. An association rule (AR) is an implication of the form $ A \Rightarrow C $, which can be read as ``if antecedent $A$ is true then consequent $C$ must be true'', where $A$ and $C$ are sets of different items (itemsets) in a database.
 An AR is defined by its antecedent, its consequent and two measures~\cite{geng2007choosing}.The first one is the support, which is the proportion of rows in the dataset where both the antecedent and the consequent appear. The second measure is the confidence,  the conditional probability to observe the consequent given an observation of the antecedent.

The most widely-used mining strategies Apriori~\cite{agrawal1994fast} and other exhaustive strategies~\cite{borgelt2005keeping,borgelt2005implementation,pei2007h} typically work by first mining frequent itemsets, then combining those itemsets to produce association rules.
However, all these algorithms face the same problems: the number of rules they produce increases exponentially with the number of items in the database, and thus it becomes impossible for a human to sort through the rules returned to pick out the best ones~\cite{fund2019comparing}. Their execution time also become an issue with massive datasets~\cite{fund2019comparing}. Finally, these algorithms need support and confidence thresholds in order to efficiently search through the solution space, 
and those thresholds need to be carefully chosen: low values can lead to long execution times and an overabundance of rules, while high values cause the algorithm to miss interesting rules. 

Researchers have tried to address these issues using top-$k$ algorithms~\cite{fournier2012mining}, metaheuristics~\cite{telikani2020survey} such as NSGAII~\cite{martin2014qar}, parallelism~\cite{heraguemi2016multi}, and GPU~\cite{djenouri2017gpu}. Meanwhile, deep learning has been use in research from image analysis~\cite{lore2017llnet} to natural language processing~\cite{ramponi2020neural} and have demonstrated a formidable capacity to find underlying relationships between input data \cite{lecun2015deep}. Auto-encoders (AE) are a kind of neural network often used to find clusters~\cite{min2018survey} or to learn representations \cite{zhuang2015supervised}. AE are trained in a supervised way using a dataset as an input and the same dataset as a target .  

Thus, AE seem applicable to ARM, because while trying to rebuild the original dataset, the AE will learn link between columns in the dataset, we then can use those links to mine AR. 

 The contributions of this paper are therefore twofold:
\begin{itemize}
    \item  We propose a new approach to mine AR using AE, which we call ARM-AE.\\
    \item We compare our approach to FP-Growth, an exhaustive state-of-the-art algorithm, and NSGAII, a commonly used metaheuristic, based on execution time, number of rules produced, average support and confidence of the rules produced, and coverage (the percentage of FP-Growth rules that our approach finds). We use three  categorical datasets taken from the popular UCI repository~\cite{Dua:2019}, ranging from a small dataset to a massive one. In all cases, ARM-AE is able to efficiently mine a limited number of high-quality rules, without the need to specify any mandatory thresholds.
\end{itemize}

The rest of this paper is organized as follows. In section \ref{Related Works} we present the literature related to the usage of neural networks to solve ARM problems. 
In section \ref{model} we present our ARM-AE model. We conduct experiments with it and with FP-Growth as a benchmark in \ref{Experiments}, and discuss the results in section \ref{Results}. Finally, we provide a synthesis of our results and concluding remarks in section \ref{Conclusion}.

\section{Related Work}
\label{Related Works}
Several previous work have used deep learning within the context of ARM.
For example, AR and a deep neural network have been used previously to predict car accidents based on tweets~\cite{zhang2018deep}. More specifically, AR were mined using Apriori on the tweets' tokens and used as input for a deep belief network to classify whether a tweet is about a car accident or not.
AR and convolutional neural networks (CNNs) have also been used together for classifying reviews of three Spanish monuments and extracting the aspects that visitors didn't like to help the monuments to improve the visitors' experience~\cite{valdivia2020people}. CNNs were first used to extract the aspects of monuments people felt strongly about. Then they were clustered into groups with the same meaning but represented by different words.  Finally AR were mined  to summarize  sentiment paired with  aspects.
As a last example, Apriori has been used together with deep learning to predict which anti-cancer drugs should be given to an individual patient~\cite{vougas2017deep}. In this work, AR were mined from a pharmacogenomics dataset, and the extracted features were used to train a deep neural network.

While examples of previous works cited above perform ARM and deep learning, they do not perform ARM using deep learning. Few works have tackled that challenge. One of them is  \cite{kishor2018association}, in which  a neural network and a genetic algorithm were used to mine AR. They used a self-organizing map (SOM) to mine clustered pattern from the data, and then used a genetic algorithm to create the AR based on these patterns. 
In \cite{li2021intelligent}, a convolutional neural network was used to mine AR on medical datasets with complex attributes, then a probability estimation method validate the AR found. 

None of the systems above use deep-learning to mine ARs directly, they either use deep learning to mine frequent itemsets or they use it alongside of ARM algorithms. Testing itemset combinations to find good ARs is a very time consuming task. The proposed ARM-AE approach mines ARs directly and thus doesn't spend time trying different combinations.

\section{ARM-AE: Association Rule Mining with Auto-Encoders }
\label{model}
In the ARM-AE context an item is the name of a column of a dataset and an itemset is a group of item with no duplicates.
In order to compare the different rules we use the support and the confidence. The support is define as the proportion of rows in the dataset $D$ where both $A$ and $C$ appear: $Supp(A,C) = |A \cap C|/|D|$, with $A$ the antecedent and $C$ the consequent, two itemsets without item in common. The confidence is the conditional probability to observe the consequent given an observation of the antecedent: $Conf(A,C) = |A \cap C|/|A|$.

We propose to leverage the representation power of auto-encoders (AEs) to directly extract AR.
An AE consists of two parts, first the encoder which consists of several layers of decreasing size, and second the decoder which consists of several layer of increasing size. The output of the encoder is the input of the decoder. The training goal of an AE is for the encoder to generate a compressed representation of the input, and for the decoder to generate a transformed version of the original input.
Training for ARM-AE doesn't differ from a classical AE training. The input and the requested output of the AE are each row of the dataset. The AE will thus learn to encode an input itemset in relation to similar itemsets, and the decoded output will be the similar itemsets, with item having a value between 0 and 1 representing how likely they are to belong in that itemset. Given that our goal is to learn a representation of the entire dataset, there is no need for evaluation or test datasets, and we train our model on the entire dataset. During training the AE will learn to recreate the input binary row, thus after training if an itemset  is used as input, the AE will recreate it, but not perfectly. Items which appear frequently with the input itemset in the dataset will have higher output value than items which never appear with the input itemset in the dataset.

The core of our ARM-AE approach is presented in  Algorithm \ref{algorithm:ARM-AE}. Our algorithm tries to mine a set of $N \times M$ rules, or $N$ rules of maximum antecedent length $M$ along with the most interesting subsets of the antecedent itemset, 
for each consequent. In ARM-AE, each item in the dataset becomes a consequent (line \ref{forallconsequent}), though it would be trivial to change that line to focus the algorithm on a specific subset of items of interest. The consequent is composed of only one item, which is the most common use-case for ARM. 
\begin{algorithm*}
\centering 
  \caption{Association Rule Mining Auto-Encoder.}\label{algorithm:ARM-AE}
  \begin{algorithmic}[1]
    \Function{ARM-AE}{$N, M, L$} 
      \Comment{$N$ number of maximum-length rules per consequent, 
      $M$ maximum number of antecedents per rule, $L$ similarity threshold}
      \State $\textrm{Consequents}\gets \textrm{items}$ 
      \State $\textrm{AE} \gets $trained Auto-encoder
      \State $\textrm{Rules} \gets \emptyset$
        \For{ $\forall \textrm{Consequent} \in  \textrm{Consequents}$}\label{forallconsequent}
            \State $\textrm{ConsequentRules} \gets \emptyset$
            \For{from $I=1$ to $N$}\label{maxNumberRules}
                \State $\textrm{Antecedents} \gets \emptyset$\label{onlyInput1}
                \For{from $J=1$ to $M$}\label{maxNumberAnt}
                    \State $\textrm{InputArray} \gets \textrm{Antecedents} \cup \textrm{Consequent}$ \label{inputAE}
                    \State $\textrm{AEOutput} \gets \textrm{AE.}\textit{forward}\textrm{(InputArray)}$\label{onlyInput2}
                    \State $\textrm{AEOutput} \gets \textit{Sort}\textrm{(AEOutput)}$\label{AddToRules1}
                    \For{$\textrm{Antecedent} \in \textrm{AEOutput}$}
                        \If{$\textrm{Antecedent} \neq \textrm{Consequent}$ and $ \textrm{Antecedent} \notin \textrm{Antecedents}$ and $  \textit{ComputeSimilarity}\textrm{(ConsequentRules,Antecedents} \cup \textrm{Antecedent}) \leq L$}\label{computeSimi}
                            \State $\textrm{Antecedents} \gets \textrm{Antecedents} \cup \textrm{Antecedent}$
                            \State $\textrm{ConsequentRules} \gets \textrm{ConsequentRules} \cup \textrm{Antecedents}$\label{AddToRules2}
                            \State Exit loop of line 13 (go to line 9)
                        \EndIf 
                    \EndFor
                \EndFor
                \State $\textrm{Rules} \gets \textrm{Rules} \cup \textrm{ConsequentRules}$
            \EndFor
           
      \EndFor
      \State \textbf{return} Rules 
    \EndFunction
  \end{algorithmic}
\end{algorithm*}
Initially, the consequent is the only input for the AE (lines \ref{onlyInput1}-\ref{onlyInput2}). The AE gives as output a score between 0 and 1 for each item in the dataset; the higher the score, the higher the probability that this item appears often with the AE input. This output is sorted in decreasing order, and the item with the highest value that isn't the consequent, already part of the antecedent, or would make the antecedent more similar than a threshold $L$ to a previously-discovered one as describe in paragraph \ref{similarity}, is added to the antecedent of the rule and this new rule is added to the set of AR for that consequent (lines \ref{AddToRules1}-\ref{AddToRules2}). The antecedent is appended to the consequent as input to the AE (line \ref{inputAE}) and the loop repeats, adding one item at a time to the antecedent until it reaches the maximum length of $M$ items (line \ref{maxNumberAnt}). Then another rule is generated from scratch for that consequent until $N$ rules are found (line \ref{maxNumberRules}) before moving on to the next consequent (line \ref{forallconsequent}).The complete code of ARM-AE is available on GitHub.\footnote{https://github.com/TheophileBERTELOOT/ARM-AE}.


\subsection{AE Training and Hyperparameters}
As was shown previously, there are three main hyperparameters to our algorithm, namely the number of maximum-length rules produced for each consequent $N$, the maximum number of items in the antecedent $M$, and the maximum similarity threshold $L$. 
There are also the hyperparameters for the AE itself, such as the number of layers and of neurons per layer, the number of training epochs, the learning rate, the activation function, the loss function, and the optimizer. These are typical for all AE architectures.

\subsection{Similarity}
\label{similarity}
An important step is computing the similarity between a new antecedent and those already found for a consequent, at line \ref{computeSimi} of Algorithm \ref{algorithm:ARM-AE}. Without this step, starting from the same consequent, the AE would always discover the same antecedent. Imposing a maximum level of similarity between a new antecedent and those already discovered forces the AE to explore the solution space and to generate a variety of different AR for each consequent.  

For our implementation of ARM-AE, the similarity metric we use is the maximum overlap between the items in the antecedent of the candidate rule and of existing rules for the same consequent, expressed as a percentage of the length of the existing rule. We only consider the similarity to rules of the same length as the candidate or longer, since otherwise an existing short and general rule would prevent the discovery of longer and more specific candidate rules. We present our similarity function in Algorithm \ref{algorithm:similarity}. We can note however that this similarity function is not necessary for ARM-AE to function; another problem-specific similarity function could be used instead.



\begin{algorithm*}
\centering 
   \caption{Compute similarity between a set of antecedents and a candidate antecedent}\label{algorithm:similarity}
   \begin{algorithmic}[1]
       \Function{Compute Similarity}{$\textrm{PreviousAntecedents,Candidate}$} \\
      \Comment{$\textrm{PreviousAntecedents}$ the list of antecedents find for a specific consequent, 
      $\textrm{Candidate}$ the antecedent whose similarity is being calculated}
      \State $\textrm{MaximumSimilarity} \gets 0$
      \For{$\forall \textrm{Antecedent} \in \textrm{PreviousAntecedents}$}
        \If { $ \textrm{|Antecedent|} <  \textrm{|Candidate|}$ }
          \State Skip Antecedent
        \EndIf
        \State $\textrm{Similarity} \gets 0$
        \For{$\forall Item \in \textrm{Antecedent}$}
            \If{$\textrm{Item} \in \textrm{Candidate}$}
               $ \textrm{Similarity} \gets  \textrm{Similarity}  + 1$
            \EndIf
        \EndFor
        \State $\textrm{Similarity} \gets \frac{\textrm{Similarity}}{\textrm{|Antecedent|}}$
        \State $\textrm{MaximumSimilarity} \gets \textit{max}\textrm{(MaximumSimilarity,similarity)}$
      \EndFor
            \State \textbf{return} $\textrm{MaximumSimilarity} $
          \EndFunction
   \end{algorithmic}
\end{algorithm*}

\section{Experiments}
\label{Experiments}
\subsection{Datasets}
We tested ARM-AE using three categorical datasets taken from the popular UCI repository~\cite{Dua:2019}. 
Those datasets were selected because they have a wide range of number of rows and items. Each dataset is preprocessed 
by replacing each categorical attribute with a one-hot set of binary attributes, with one column for each category value. 
Table~\ref{tab:datasets} presents the list of datasets along with their number of rows and attributes after binarization.
\begin{table}
\centering
\caption{UCI datasets selected for experiments.}
\label{tab:datasets}
\begin{tabular}{||c r c l l||} 
 \hline
    Dataset & Rows & Columns & Support & Confidence   \\ [0.5ex] 
 \hline\hline
   Chess & 3,196 & 75 & 0.005 & 0.01 \\
 \hline
    Nursery & 12,960 & 32 & 0.01 & 0.01 \\
 \hline
    Plants & 34,781 & 70 & 0.005 & 0.015 \\
 \hline
\end{tabular}
\vspace{0.2cm}
\end{table}

\subsection{Hyperpameters}

For the architecture of our AE, both the encoder and decoder are made up of three fully connected layers. The number of neurons for each layer is the number of different items in the dataset. This architecture has shown the best results in our experiments. 
The neurons use the tanh function shown in Equation \ref{eq:tanh} as their activation function because ARM-AE is build to approximate the support of the rules and  support is define between 0 and 1 therefore we choose a common activation function define between 0 and 1. The learning rate is $10^{-3}$.  The MSELoss shown in equation \ref{eq:loss} is our loss function (where $x$ is the input and $y$ the target), and we use the Adam optimizer. 

\begin{equation}
Tanh(x) = \frac{\exp{x}-\exp{-x}}{\exp{x}+\exp{-x}}
\label{eq:tanh}
\end{equation}
\begin{equation}
MSELoss(x,y) = (x-y)^{2}
\label{eq:loss}
\end{equation}

\begin{figure}
\centering
   {\includegraphics[width=4in]{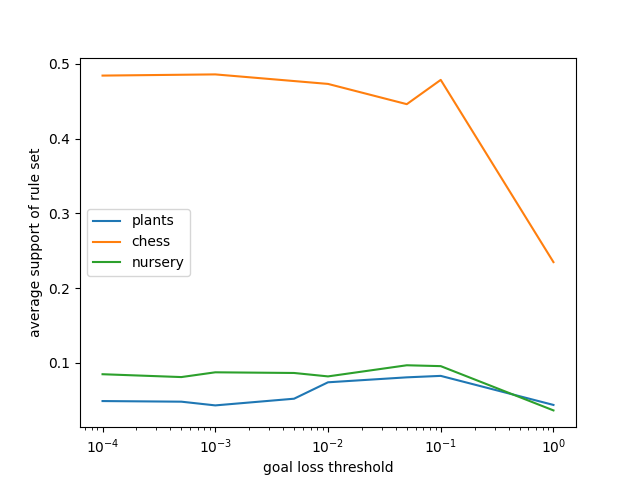}}
    \caption{Average support of the ARM-AE rule set for the three datasets using different thresholdd values of goal loss.}
    \label{fig:goalLosses}
\end{figure}

We set the algorithm to look for rules with a maximum of two items in the antecedent ($M = 2$) and for two rules per consequent ($N = 2$). The similarity threshold $L = 0.5$, so there will be a maximum of one item overlap between the antecedents of the two pairs of rules. In order to set the number of training epochs, we use the difference of loss between two epochs; when this difference becomes lower than a threshold  then we take the previous epoch as our final model. We tested several values of for this threshold. Figure \ref{fig:goalLosses} shows the average support of the rules found by ARM-AE for different values of the threshold on the three datasets; we did 10 experiments for each threshold value and trained for 100 epochs in each case, and show the average in the figure. 
A threshold of 1 mean that there is no training at all, and the result is the a random initialisation of the network. The figure shows that the best threshold is 0.1, which in practice mean 1 to 5 epoch. Performance decreases significantly after this threshold. Thus we choose to use 0.1 as our threshold for the rest of our experiments.

\subsection{Benchmarks}
We use FPGrowth~\cite{borgelt2005keeping} as one of our two benchmarks, as it is one of the fastest and most popular state-of-the-art exhaustive ARM algorithms. We used the implementation found in the Mlxtend library \cite{raschkas_2018_mlxtend}. The support and confidence thresholds for each dataset are picked so the number of rules mined with one or two items in the antecedents and one item in the consequent is nearly the same as the number of rules produced by our algorithm. The thresholds are included in Table \ref{tab:datasets}. Our second benchmark is the   NSGAII~\cite{deb2002fast} algorithm, chosen because genetic algorithms and domination concepts are a dominant trend in ARM \cite{telikani2020survey} and NSGAII is a commonly-used baseline for meta-heuristic. We used the DEAP library \cite{fortin2012deap} to implement our NSGAII. 
For our benchmark algorithm NSGAII, we used a population of 100 individuals, a crossover ratio of 0.9 and a mutation probability of 0.01. We set a maximum of 2 antecedents and 1 consequent in the rules. We keep the 300 best rules found across all generation for the plants and chess datasets and the 150 best for the nursery dataset, so results are comparable with ARM-AE and FP-Growth outputs. To set the number of generation we check if the average support and average confidence of the best rules found so far increase above than a threshold. If both decrease or increase less than the threshold then the NSGAII algorithm stops. We use a threshold of 0.01.

\subsection{Metrics}
We compute several metrics to compare ARM-AE and our benchmarks. The first one is the number of rules found by ARM-AE that have a support greater than 0. A rule that has a support greater than 0 is one that actually exists in the dataset. But since ARM-AE does not compute the support of the rules during its search, unlike FP-Growth and other exhaustive algorithms, it offers no guarantee in that respect, thus the importance of this metric. 
The second metric is the percentage of rules found by FP-Growth that are also found by our algorithm and NSGAII. Since FP-Growth performs an exhaustive search for all the best rules, this will indicate how close to complete the ARM-AE and NSGAII rule sets are. 
Recall however that we set ARM-AE's hyperparameters to generate four AR per consequent with only one or two items in the antecedent.
To have a fair comparison, we have therefore set FP-Growth's thresholds to extract the best four AR per consequent with that length, as mentioned previously. Next, we compare the execution times for all three algorithms. 
Finally we compare the average support and confidence of the rules produced by the three algorithms.

\section{Results and Analysis}
\label{Results}

We trained and tested our AE and NSGAII independently 10 times with each dataset, in order to account for differences in the network's random initialization and the stochastic nature of meta-heuristics. We report the average results over all 10 runs. The FPGrowth algorithm, having no random initialization, was only run once.

The first metric to consider is the percentage of rules produced by ARM-AE that have a support greater than 0. The results shown in Table \ref{tab:supp>0} indicate that up to 2\% of the rules produced actually don't exist in the dataset. Since ARM-AE is required to generate a number of rules for each consequent, some very-low-support or zero-support rules will be included when no high-support rules exist for a consequent. Without a computation of support, ARM-AE cannot detect or filter out these cases. This is a limitation of our approach. However, since such rules account for less than 2\% of the extracted rule set, the amount of noise included is not a major issue.

\begin{table}
\centering
\caption{Percentage of rules with support greater than 0}
\label{tab:supp>0}
\begin{tabular}{||c c ||} 
 \hline
    Dataset & Rules \\ [0.5ex] 
 \hline\hline
   Chess & 0.99 \\ 
 \hline
    Nursery & 0.98 \\ 
 \hline
    Plants & 1.00  \\ 
 \hline
\end{tabular}
\vspace{0.2cm}
\end{table}

The next metrics compare the AR set mined by ARM-AE, NSGAII and FP-Growth. First we consider the proportion of FP-Growth's rule set that is mined by ARM-AE and NSGAII. FP-Growth exhaustively mines all rules with one consequent and one or two antecedents that are above the set support and confidence threshold. For its part, ARM-AE discovers between 20\% and 44\% of that set, as is shown in Table \ref{tab:avg}. 
Regarding NSGAII, it mines up 8\% of the rules set mined by FP-Growth. That can be explain by the highly stochastic aspect of genetic algorithm and because NSGAII focuses on non-dominated rules where FP-Growth  focuses on high support and high confidence rules. In non-dominated mining, if a rule has a very low support but a very high confidence and no other rules have a higher support with at least the same confidence, then it may not be kept.
Table \ref{tab:avg} also shows that the average support and confidence of the rules mined by our three algorithms. The results show that the rules found by ARM-AE have a support and confidence almost as high as those mined by FP-Growth, and in fact the confidence is higher in the case of the Plants dataset. That is despite ARM-AE including up to 2\% of zero-support rules in its results, as mentioned previously. This shows that, although ARM-AE mines less than half the rules found by FP-Growth, it mines a high-quality subset of the rules that exist in the database. NSGAII mines a rule set with very high confidence and lower support than the two other algorithms. ARM-AE mines a rule set that is more similar to the one found by FP-Growth than NSGAII.

\begin{table*}
\centering
\caption{Coverage, support and confidence results per algorithm and dataset.}
\begin{tabular}{||c| c c c  | c c c  | c c ||} 
 \hline
\multirow{1}{*}{Dataset} &
\multicolumn{3}{c|}{ ARM-AE }&
\multicolumn{3}{c|}{ NSGAII }&
\multicolumn{2}{c||}{FP-Growth}\\
   & coverage & support & confidence& coverage & support & confidence & support & confidence \\ [0.5ex] 
 \hline\hline
   Chess & 0.20 & 0.48 & 0.50 & 0.02 & 0.33 & 0.71 & 0.51 & 0.54 \\ 
 \hline
    Nursery & 0.40 & 0.1 & 0.33 & 0.08 & 0.06 & 0.38 & 0.12 & 0.37\\ 
     \hline
    Plants & 0.44 & 0.08 &  0.75 & 0.02 & 0.05 & 0.78 & 0.09 & 0.7 \\ 
 \hline
\end{tabular}
\vspace{0.2cm}
\label{tab:avg}
\end{table*}

Finally, we compare the performance of the three algorithms in terms of execution time and total number of rules generated, in Table \ref{tab:performance}. We consider first the execution time for the entire algorithm, including training the AE in the case of ARM-AE. The most time-consuming step in every ARM algorithm is to compute the support and confidence of rules. FP-Growth use a Fp-tree to speed up the mining of frequent itemsets, but then must compute the support and confidence of all the pairs of itemsets. NSGAII compute the support and confidence for a large part of the population at each generation explaining why it is slower than FP-Growth. By contrast, ARM-AE compute the support and confidence only for the subset of rules mined. For the three datasets, ARM-AE is much more efficient than FP-Growth and NSGAII in terms of execution time. 
In terms of number of rules, ARM-AE and NSGAII have the advantage of generating a small rule set whose size is controlled by the algorithm's hyperparameters. By contrast, the exhaustive search of FP-Growth generates tens to hundreds of thousands of rules. As mentioned previously, this is one of the main weaknesses of algorithms such as this one  \cite{fund2019comparing}: it is impossible for a human to read and understand such a massive set of rules, and additional filtering or ranking algorithms must then be applied in post-processing before the rules are suitable for human use. ARM-AE and NSGAII do not suffer from that issue. 

\begin{table*}
\centering
\caption{Performance of the Algorithms}
\label{tab:performance}
\begin{tabular}{||c| c c| c c| c c||} 
 \hline
\multirow{1}{*}{Dataset} &
\multicolumn{2}{c|}{ ARM-AE }&
\multicolumn{2}{c|}{ NSGAII }&
\multicolumn{2}{c||}{FP-Growth}\\
     & Execution (s) & Rules& Execution (s) & Rules & Execution (s) & Rules \\ [0.5ex] 
 \hline\hline
   Chess & 1.2 & 296& 95.0 & 241 & 35.2 & 278,823\\ 
 \hline
    Nursery & 0.8 & 124& 99.0 & 150 & 1.4 & 18,140\\ 
     \hline
    Plants & 1.8 & 280& 979.0 & 261 & 114 & 269,586\\ 
 \hline
\end{tabular}
\vspace{0.2cm}
\end{table*}




\section{Conclusion}
\label{Conclusion}
In this paper, we introduced a new deep learning architecture to mine association rules, the Association Rules Mining Auto-Encoder(ARM-AE) algorithm. 
We compared this new algorithm to the popular state-of-the-art FP-Growth algorithm and to the state-of-the-art genetic algorithm NSGAII on three categorical datasets. 
Our results show that ARM-AE produces a set of rules with an average support and confidence close to that of FP-Growth and with higher support than the one of NSGAII, but with a much smaller execution time and by generating a much smaller number of rules than FP-Growth, which makes the results much more usable for a human being. 
Moreover, ARM-AE has hyperparameters to control the number of rule produced for each consequent, the maximum number of items in the antecedent of the rules, and the similarity between the rules, which again makes the algorithm much more user-friendly. 
The main drawbacks of ARM-AE stem from the fact it does not compute support and confidence for its rules, which mean it cannot filter out zero-support rules and cannot give an importance ranking of the rules it mined, but these will be addressed in future works.

\section*{Acknowledgement}
This research was made possible by the support of the INSPQ, as well as the financial support of the Canadian research funding agencies CIHR and NSERC.

\bibliographystyle{unsrt}  
\bibliography{references}

\begin{thebibliography}{10}

\bibitem{agrawal1994fast}
Rakesh Agrawal, Ramakrishnan Srikant, et~al.
\newblock Fast algorithms for mining association rules.
\newblock In {\em Proc. 20th int. conf. very large data bases, VLDB}, volume
  1215, pages 487--499. Citeseer, 1994.

\bibitem{nair2015stock}
Binoy~B Nair, VP~Mohandas, Nikhil Nayanar, ESR Teja, S~Vigneshwari, and KVNS
  Teja.
\newblock A stock trading recommender system based on temporal association rule
  mining.
\newblock {\em SAGE Open}, 5(2):2158244015579941, 2015.

\bibitem{alkeshuosh2017using}
Azhar~Hussein Alkeshuosh, Mariam~Zomorodi Moghadam, Inas Al~Mansoori, and
  Moloud Abdar.
\newblock Using pso algorithm for producing best rules in diagnosis of heart
  disease.
\newblock In {\em 2017 international conference on computer and applications
  (ICCA)}, pages 306--311. IEEE, 2017.

\bibitem{geng2007choosing}
Liqiang Geng and Howard~J Hamilton.
\newblock Choosing the right lens: Finding what is interesting in data mining.
\newblock In {\em Quality measures in data mining}, pages 3--24. Springer,
  2007.

\bibitem{borgelt2005keeping}
Christian Borgelt.
\newblock Keeping things simple: finding frequent item sets by recursive
  elimination.
\newblock In {\em Proceedings of the 1st international workshop on open source
  data mining: frequent pattern mining implementations}, pages 66--70, 2005.

\bibitem{borgelt2005implementation}
Borgelt Christian.
\newblock An implementation of the fp-growth algorithm.
\newblock In {\em Proceedings of the 1st international workshop on open source
  data mining: frequent pattern mining implementations}, pages 1--5, 2005.

\bibitem{pei2007h}
Jian Pei, Jiawei Han, Hongjun Lu, Shojiro Nishio, Shiwei Tang, and Dongqing
  Yang.
\newblock H-mine: Fast and space-preserving frequent pattern mining in large
  databases.
\newblock {\em IIE transactions}, 39(6):593--605, 2007.

\bibitem{fund2019comparing}
Ian Fund.
\newblock {\em Comparing Association Rules and Deep Neural Networks on Medical
  Data}.
\newblock PhD thesis, University of Houston, 2019.

\bibitem{fournier2012mining}
Philippe Fournier-Viger, Cheng-Wei Wu, and Vincent~S Tseng.
\newblock Mining top-k association rules.
\newblock In {\em Canadian Conference on Artificial Intelligence}, pages
  61--73. Springer, 2012.

\bibitem{telikani2020survey}
Akbar Telikani, Amir~H Gandomi, and Asadollah Shahbahrami.
\newblock A survey of evolutionary computation for association rule mining.
\newblock {\em Information Sciences}, 524:318--352, 2020.

\bibitem{martin2014qar}
D~Mart{\'\i}n, Alejandro Rosete, Jes{\'u}s Alcal{\'a}-Fdez, and Francisco
  Herrera.
\newblock Qar-cip-nsga-ii: A new multi-objective evolutionary algorithm to mine
  quantitative association rules.
\newblock {\em Information Sciences}, 258:1--28, 2014.

\bibitem{heraguemi2016multi}
Kamel~Eddine Heraguemi, Nadjet Kamel, and Habiba Drias.
\newblock Multi-swarm bat algorithm for association rule mining using multiple
  cooperative strategies.
\newblock {\em Applied Intelligence}, 45(4):1021--1033, 2016.

\bibitem{djenouri2017gpu}
Youcef Djenouri, Ahcene Bendjoudi, Djamel Djenouri, and Marco Comuzzi.
\newblock Gpu-based bio-inspired model for solving association rules mining
  problem.
\newblock In {\em 2017 25th Euromicro International Conference on Parallel,
  Distributed and Network-Based Processing (PDP)}, pages 262--269. IEEE, 2017.

\bibitem{lore2017llnet}
Kin~Gwn Lore, Adedotun Akintayo, and Soumik Sarkar.
\newblock Llnet: A deep autoencoder approach to natural low-light image
  enhancement.
\newblock {\em Pattern Recognition}, 61:650--662, 2017.

\bibitem{ramponi2020neural}
Alan Ramponi and Barbara Plank.
\newblock Neural unsupervised domain adaptation in nlp---a survey.
\newblock {\em arXiv preprint arXiv:2006.00632}, 2020.

\bibitem{lecun2015deep}
Yann LeCun, Yoshua Bengio, and Geoffrey Hinton.
\newblock Deep learning.
\newblock {\em nature}, 521(7553):436--444, 2015.

\bibitem{min2018survey}
Erxue Min, Xifeng Guo, Qiang Liu, Gen Zhang, Jianjing Cui, and Jun Long.
\newblock A survey of clustering with deep learning: From the perspective of
  network architecture.
\newblock {\em IEEE Access}, 6:39501--39514, 2018.

\bibitem{zhuang2015supervised}
Fuzhen Zhuang, Xiaohu Cheng, Ping Luo, Sinno~Jialin Pan, and Qing He.
\newblock Supervised representation learning: Transfer learning with deep
  autoencoders.
\newblock In {\em Twenty-Fourth International Joint Conference on Artificial
  Intelligence}, 2015.

\bibitem{Dua:2019}
Dheeru Dua and Casey Graff.
\newblock {UCI} machine learning repository, 2017.

\bibitem{zhang2018deep}
Zhenhua Zhang, Qing He, Jing Gao, and Ming Ni.
\newblock A deep learning approach for detecting traffic accidents from social
  media data.
\newblock {\em Transportation research part C: emerging technologies},
  86:580--596, 2018.

\bibitem{valdivia2020people}
Ana Valdivia, Eugenio Mart{\'\i}nez-C{\'a}mara, Iti Chaturvedi, M~Luzon, Erik
  Cambria, Yew-Soon Ong, and Francisco Herrera.
\newblock What do people think about this monument? understanding negative
  reviews via deep learning, clustering and descriptive rules.
\newblock {\em Journal of Ambient Intelligence and Humanized Computing},
  11(1):39--52, 2020.

\bibitem{vougas2017deep}
Konstantinos Vougas, Magdalena Krochmal, Thomas Jackson, Alexander Polyzos,
  Archimides Aggelopoulos, Ioannis~S Pateras, Michael Liontos, Anastasia
  Varvarigou, Elizabeth~O Johnson, Vassilis Georgoulias, et~al.
\newblock Deep learning and association rule mining for predicting drug
  response in cancer. a personalised medicine approach.
\newblock {\em BioRxiv}, page 070490, 2017.

\bibitem{kishor2018association}
Peddi Kishor and Porika Sammulal.
\newblock Association rule mining using an unsupervised neural network with an
  optimized genetic algorithm.
\newblock In {\em International Conference on Communications and Cyber Physical
  Engineering 2018}, pages 657--669. Springer, 2018.

\bibitem{li2021intelligent}
Xiaofeng Li, Dong Li, Yuanbei Deng, and Jinming Xing.
\newblock Intelligent mining algorithm for complex medical data based on deep
  learning.
\newblock {\em Journal of Ambient Intelligence and Humanized Computing},
  12(2):1667--1678, 2021.

\bibitem{raschkas_2018_mlxtend}
Sebastian Raschka.
\newblock Mlxtend: Providing machine learning and data science utilities and
  extensions to python’s scientific computing stack.
\newblock {\em The Journal of Open Source Software}, 3(24), April 2018.

\bibitem{deb2002fast}
Kalyanmoy Deb, Amrit Pratap, Sameer Agarwal, and TAMT Meyarivan.
\newblock A fast and elitist multiobjective genetic algorithm: Nsga-ii.
\newblock {\em IEEE transactions on evolutionary computation}, 6(2):182--197,
  2002.

\bibitem{fortin2012deap}
F{\'e}lix-Antoine Fortin, Fran{\c{c}}ois-Michel De~Rainville,
  Marc-Andr{\'e}~Gardner Gardner, Marc Parizeau, and Christian Gagn{\'e}.
\newblock Deap: Evolutionary algorithms made easy.
\newblock {\em The Journal of Machine Learning Research}, 13(1):2171--2175,
  2012.

\end{thebibliography}

\end{document}